\documentclass[a4paper]{article}

\usepackage{geometry}
\usepackage{url}
\usepackage{hyperref}
\usepackage{tikz}
\usetikzlibrary{trees}
\usepackage{fancyvrb}
\usepackage{fancyhdr}
\usepackage{xcolor}
\hypersetup{
   colorlinks,
   citecolor=blue,
   filecolor=blue,
   linkcolor=blue,
   urlcolor=blue
}

\newcommand\blankfootnote[1]{%
  \let\thefootnote\relax\footnotetext{#1}%
  \let\thefootnote\svthefootnote%
}

\begin{document}

\title{PonyGE2: Grammatical Evolution in Python}

\maketitle

\begin{center}

\author{Michael~Fenton\footnote{\label{NCRA}Michael~Fenton, James~McDermott, David~Fagan, Stefan~Forstenlechner, and Michael~O'Neill are with the Natural Computing Research and Applications group (NCRA) in UCD, Ireland (e-mail: \href{mailto:michaelfenton1@gmail.com}{michaelfenton1@gmail.com}, \href{mailto:james.mcdermott2@ucd.ie}{james.mcdermott2@ucd.ie}, \href{mailto:david.fagan@ucd.ie}{david.fagan@ucd.ie}, \href{mailto:stefan.forstenlechner@ucdconnect.ie}{stefan.forstenlechner@ucdconnect.ie}, \href{mailto:m.oneill@ucd.ie}{m.oneill@ucd.ie}).}, James~McDermott\textsuperscript{\ref{NCRA}}, David~Fagan\textsuperscript{\ref{NCRA}}, Stefan~Forstenlechner\textsuperscript{\ref{NCRA}}, Erik~Hemberg\footnote{Erik~Hemberg is with the Computer Science and Artificial Intelligence Labratory (CSAIL) in MIT (email: \href{mailto:erik.hemberg@gmail.com}{erik.hemberg@gmail.com}.)}, Michael~O'Neill\textsuperscript{\ref{NCRA}}}

\blankfootnote{\copyright Fenton \emph{et al.}, 2017. This is the author's version of the work. It is posted here for your personal use. Not for redistribution. The definitive version is published in Proceedings of GECCO '17 Companion, \url{http://dx.doi.org/10.1145/3067695.3082469}}

\blankfootnote{\textbf{Reference format:} Michael Fenton, James McDermott, David Fagan, Stefan Forstenlechner, Erik Hemberg, Michael O’Neill. 2017. PonyGE2: Grammatical Evolution
in Python. In \emph{Proceedings of GECCO ’17 Companion}, Berlin, Germany, July 15-19, 2017, 8 pages. DOI: \url{http://dx.doi.org/10.1145/3067695.3082469}}

\end{center}

\addtocounter{footnote}{-2}

\begin{abstract}
\label{abstract}

Grammatical Evolution (GE) is a population-based evolutionary algorithm, where a formal grammar is used in the genotype to phenotype mapping process. PonyGE2 is an open source implementation of GE in Python, developed at UCD's Natural Computing Research and Applications group. It is intended as an advertisement and a starting-point for those new to GE, a reference for students and researchers, a rapid-prototyping medium for our own experiments, and a Python workout. As well as providing the characteristic genotype to phenotype mapping of GE, a search algorithm engine is also provided. A number of sample problems and tutorials on how to use and adapt PonyGE2 have been developed.

\end{abstract}

\section{Introduction}
\label{sec:intro}

Grammatical Evolution (GE) is a grammar-based form of Genetic Programming~\cite{Koza:1992:GP}, where a formal grammar is used in the genotype to phenotype mapping process~\cite{ONeill:2003:GE}. Whereas previous releases of Grammatical Evolution have been written in C~\cite{libGE}, Java~\cite{ONeill:2008:GEVA}, R~\cite{Noorian:2015:GE_R}, and even Ruby~\cite{GERET}, PonyGE2 is an implementation of GE in Python. The original version of PonyGE~\cite{PonyGE} was designed to be short and contained in a single file. However, over time it grew to become unwieldy and a more structured approach was needed. This has led to the development of PonyGE2, presented here. PonyGE2 is intended as an advertisement and a starting-point for those new to GE, a reference for students and researchers, a rapid-prototyping medium for our own experiments, and a Python workout.

Grammatical Evolution marries principles from molecular biology to the representational power of formal grammars~\cite{ONeill:2003:GE}. GE’s rich modularity gives a unique flexibility, making it possible to use alternative search strategies, whether evolutionary, deterministic or some other approach, and to radically change its behaviour by merely changing the grammar supplied. As a grammar is used to describe the structures that are generated by GE, it is trivial to modify the output structures by editing the grammar, typically represented in plain text BNF (Backus-Naur Form) format. This is one of the main advantages that makes the GE approach so attractive. The genotype-phenotype mapping also means that instead of operating exclusively on solution trees, as in standard GP, GE allows search operators to act on the genotypes (i.e.~integer or binary lists), on partially derived phenotypes, or on the fully-formed phenotypic derivation trees themselves.

The rest of this paper is structured as follows. Section~\ref{sec:ponyge2} frames PonyGE2 against the backdrop of previous GE releases, and outlines its modular structure. Section~\ref{sec:grammars} gives an overview of grammars under PonyGE2, including how grammars are parsed using Regular Expressions in Section~\ref{subsec:parsing_grammars}, and PonyGE2's handling of special grammar characters in Section~\ref{subsec:variable_grammar_ranges}. Section~\ref{sec:linear_representation} details the linear representation of PonyGE2 (including mapping, wrapping, invalid individuals, and unit productions), while Section~\ref{sec:tree_representation} details derivation tree representations. Operators are listed in Section~\ref{sec:operators}. A list of example problems provided with PonyGE2 is given in Section~\ref{sec:examples}, before conclusions are drawn and avenues for future work identified in Section \ref{sec:future_work}.

\section{PonyGE2}
\label{sec:ponyge2}

GEVA \cite{ONeill:2008:GEVA} represented a feature-rich, mature representation of linear GE. However, the codebase was verbose and difficult to maintain or modify, and the release cycle of GEVA had stagnated due to a knowledge gap within the development community. Furthermore, advances in Java 7 and 8 were not being taken advantage of.

Python has become a widely used language, and has seen broad adoption from people with little or no programming background in both academia and industry as it provides an easy first step into data science and machine learning. Since GEVA had become verbose, the original version of PonyGE~\cite{PonyGE} was developed as a clean, compact, and overall user-friendly implementation for a user base of varying research needs and backgrounds. Recently PonyGE had seen an uptake in new users, and feedback was that while PonyGE presented a usable Python implementation of GE, the code base had become disorganised. While the original incarnation was intended to be small and compact (`pony-sized') and as such was implemented as a single source file, the continual extension of this original code base to accommodate varying requirements of different researchers negated this original goal. What was once small and compact had become large and unmanageable.

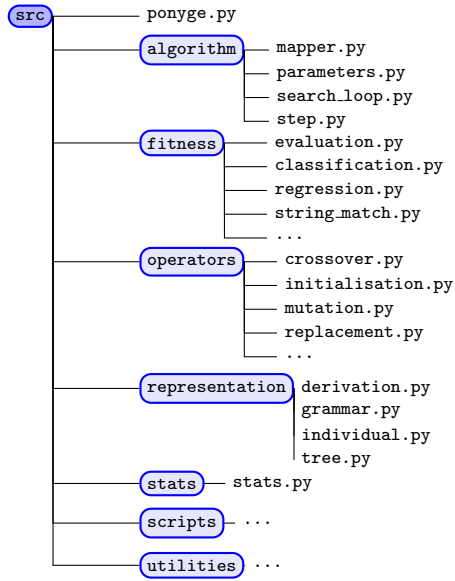
\begin{figure}
\centering
\tikzstyle{every node}=[thick,anchor=west, rounded corners, font={\scriptsize\ttfamily}, inner sep=2.5pt]
\tikzstyle{selected}=[draw=blue,fill=blue!10]
\tikzstyle{root}=[selected, fill=blue!30]

\begin{tikzpicture}[%
    scale=.7,
    grow via three points={one child at (1,0) and 
    two children at (1.15,0) and (1.15,-0.45)}, 
    edge from parent path={(\tikzparentnode.east) |- (\tikzchildnode.west)}]
  \node [root] {src}
  	child {node {ponyge.py}}
    child { node at (0, -0.19) [selected] {algorithm}
    	child { node {mapper.py}}
        child { node {parameters.py}}
        child { node {search\_loop.py}}
        child { node {step.py}}
    }   
    child { node at (0,-1.5) [selected] {fitness}
        child { node {evaluation.py}}
        child { node {classification.py}}
        child { node {regression.py}}
        child { node {string\_match.py}}
        child { node {\dots}}
    }
    child { node at (0,-3.3) [selected] {operators}
        child { node {crossover.py}}
        child { node {initialisation.py}}
        child { node {mutation.py}}
        child { node {replacement.py}}
        child { node {\dots}}
    }
    child { node at (0,-5.25) [selected] {representation}
        child { node {derivation.py}}
        child { node {grammar.py}}
        child { node {individual.py}}
        child { node {tree.py}}
    }
    child { node at (0,-6.6) [selected] {stats}
        child { node {stats.py}}
    }
    child { node at (0,-6.9) [selected] {scripts}
        child { node {\dots}}
    }
    child { node at (0,-7.25) [selected] {utilities}
    	child { node {\dots}}
    };
\end{tikzpicture}
\caption{Organizational structure of the PonyGE2 Codebase.}
\label{fig:code_base}
\end{figure}

The decision was made to merge the feature-rich and modular aspects of GEVA with Python, and to re-structure the development code base of PonyGE into a package structure. As such, the original PonyGE file was re-factored, re-written, and greatly extended to present a cleaner and simpler structure with much added functionality. This modular code base allows users to work on a single package without having to wade through thousands of lines of potentially irrelevant code. As shown in Fig.~\ref{fig:code_base}, each element of the algorithm has been confined in a modular way and the code adapted to allow for usage of multiple search engines and operators. This move harks back to some of the design choices made for GEVA~\cite{ONeill:2008:GEVA}, but also embraces the original ideology behind GE~\cite{ONeill:2003:GE,libGE}.

\begin{figure}[!ht]
    \begin{center}
    \includegraphics[width=0.44\textwidth]{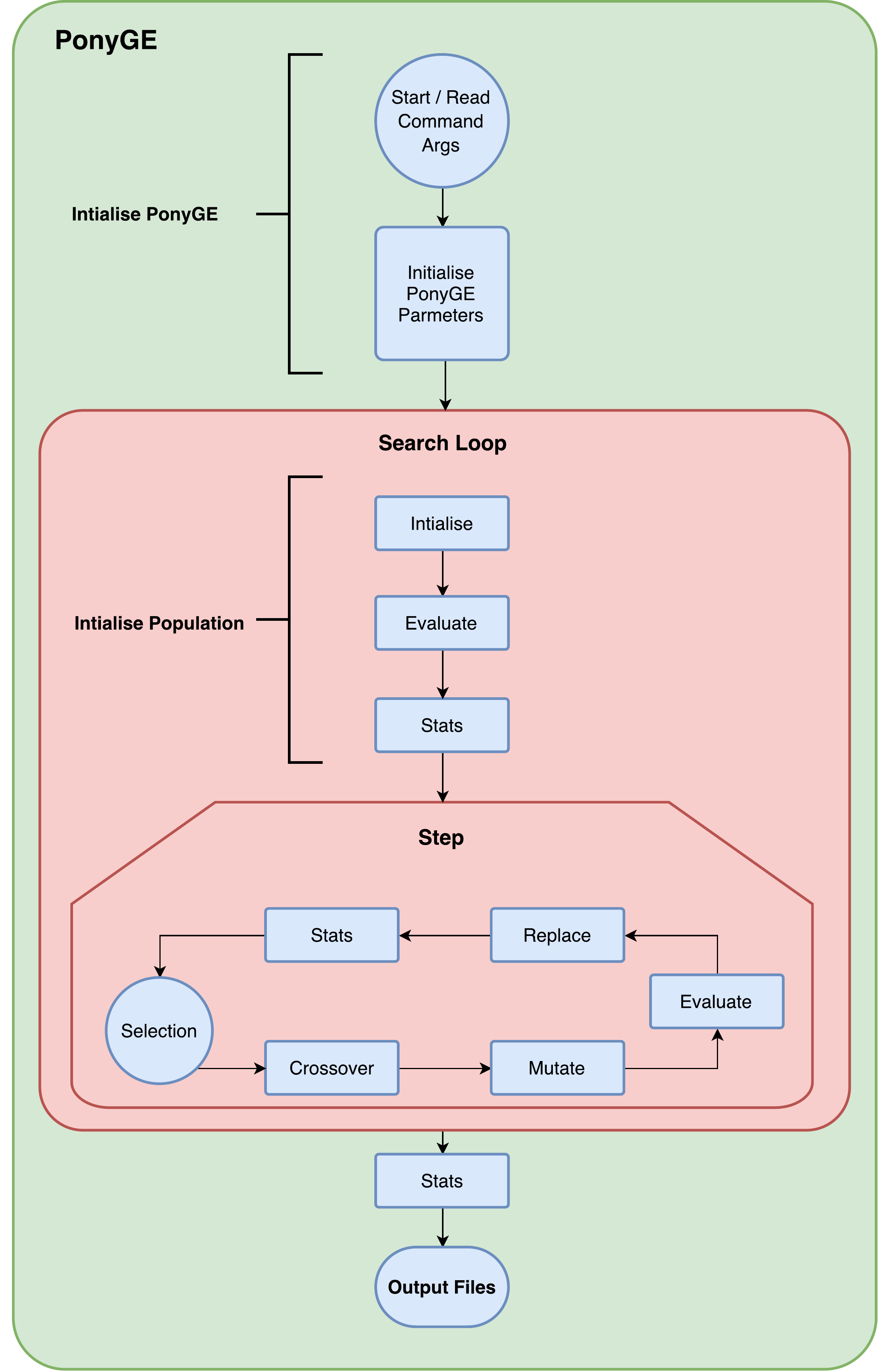}	 
    \caption{PonyGE2 control flow diagram for typical GE/GP setup.}
    \label{fig:control_flow}
  \end{center}
\end{figure}

The modular structure of PonyGE2, as shown in Fig.~\ref{fig:code_base}, allows for a high degree of flexibility in the algorithm. The control flow for a typical PonyGE2 setup is shown in Fig.~\ref{fig:control_flow}. All function blocks in Fig.~\ref{fig:control_flow} represent parametrisable functions. This means that in PonyGE2 not only is it possible to specify unique operators, but it is also possible to easily define unique step and search loop control flows. Unlike with previous official releases of GE systems which required compiling (such as C \cite{libGE} or Java \cite{ONeill:2008:GEVA}), the plug-and-play nature of Python programming coupled with the modularity of the control flow makes PonyGE2 an intuitive, highly user-friendly system that has been designed first and foremost with customisation and personalisation in mind. Furthermore, PonyGE2 is fully PEP-8 compliant \cite{PEP-8}.

A major strength of PonyGE2 is the ability to mix and match representation types. Both linear genome representations \cite{ONeill:2003:GE} and derivation tree representations \cite{Whigham:1995:GBGP} are implemented simultaneously in PonyGE2, meaning that every individual has both a genome and full derivation tree. Operators of either type can be mixed and used freely, while maintaining full compatibility with both representation types. There are advantages and disadvantages to both types, discussed later in Sections \ref{sec:linear_representation} and \ref{sec:tree_representation}.

PonyGE2 is run from the command line from within the source directory. Executing the main \texttt{ponyge.py} file will run an example regression problem\footnote{Note that the default settings do not necessarily represent suggested good settings, but are to serve primarily as examples of how to use the system.} and generate a results folder. Each results folder generated by an evolutionary run contains several files, detailing all statistics gathered over the course of the run, a graph of the best fitness plotted against generations, a documented list of all the parameters used, as well as a file detailing the best individual. An array of command line arguments are available for specifying desired parameters, which can also be specified in an external parameters file.

An important issue for any scientific field is experimental clarity and comparability, i.e.~allowing for experiments to be easily reproduced. To that extent, it is possible to exactly recreate a PonyGE2 run by using the parameters file saved from that run. Parameters files are saved automatically for each run, and include all necessary information (including random seeds) to set the parameters of a new run in order to perfectly reproduce a given experiment\footnote{Note that this is contingent on the use of the original grammar, fitness function, and datasets (if used). Note also that changes to the code may affect result outcomes.}. Furthermore, PonyGE2 comes pre-packaged with a number of benchmark datasets and grammars which can be used to verify and test previous results~\cite{McDermott:2012:Benchmarks,Nicolau:2015:Guidelines}.

The PonyGE2 project uses GitHub~\cite{PonyGE2} to allow for open usage of the code with forking and version control. This allows users to stay up to date with current releases as new functionality is rolled out. The use of GitHub also provides issue tracking and a forum for users to voice their desires/problems with the software. 

PonyGE2 requires Python 3.5 or higher, and uses the matplotlib, numpy, scipy, scikit-learn (sklearn), and pandas packages. All requirements can be satisfied with Anaconda. PonyGE2 v0.1.0 has been released under GNU GPL version 3 \cite{PonyGE2}.

\subsection{Scripts and Utilities}

Besides the main \texttt{ponyge.py} file that can be found in the \texttt{src} directory, a number of extra scripts are provided with PonyGE2. These are located in the \texttt{scripts} folder. These extra scripts have been designed to work either as standalone files, or to work in tandem with PonyGE2. Various functions from within these scripts can provide extra functionality to PonyGE2. Most prominent of the scripts are a basic experiment manager and statistics parser for executing multiple experimental runs. A full breakdown of all scripts is provided in the \texttt{README} file \cite{PonyGE2}.

The \texttt{utilities} folder provides an array of additional functions used by PonyGE2, such as file I/O, plotting, the command-line parser, protected mathematical operators, and error metrics.	


\section{Grammars}
\label{sec:grammars}

When tackling a problem with GE, a suitable grammar must initially be defined. The grammar can be either the specification of an entire programming language or, perhaps more usefully, a subset of a language geared towards the problem at hand.

In PonyGE2, Bacus-Naur Form (BNF) is used to describe the output language to be produced by the system. BNF is a notation for expressing a grammar in the form of production rules. BNF grammars consist of terminals, which are symbols that can appear in the language, e.g.~locally or globally defined variables, binary boolean operators \texttt{and}, \texttt{or}, \texttt{xor}, and \texttt{nand}, unary boolean operators \texttt{not}, constants, \texttt{True} and \texttt{False} etc.~and non-terminals, which can be expanded into one or more terminals and non-terminals.

A grammar is a set of production rules that defines a language. Each production rule is composed of a left-hand side (a single non-terminal), followed by the "goes-to" symbol \texttt{::=}, followed by a list of production choices separated by the "or" symbol \texttt{|}. Production choices can be composed of any combination of terminals or non-terminals. Non-terminals are enclosed by angle brackets \texttt{<>}. For example, consider the following production rule:

\begin{center}
    \texttt{<a> ::= <b>c | d}
\end{center}

In this rule, the non-terminal \texttt{<a>} maps to either the choice \texttt{<b>c} (a combination of a new non-terminal \texttt{<b>} and a terminal \texttt{c}), or a single terminal \texttt{d}.

\subsection{Recursion}
\label{sec:recursion}

One of the most powerful aspects of GE is that the representation can be variable in length. Notably, rules can be recursive (i.e.~a non-terminal production rule can contain itself as a production choice), which can allow GE to generate solutions of arbitrary size, e.g.:

\begin{center}
    \texttt{<a> ::= <a> + b | b}
\end{center}

The grammar is used in a developmental approach whereby the evolutionary process chooses the productions to be chosen at each stage of a mapping process, starting from the start symbol, until a complete program is formed. A complete program is one that is comprised solely from elements of the terminal set \texttt{T}.

In PonyGE2 the BNF definition is comprised entirely of the set of production rules, with the definition of terminals and non-terminals implicit in these rules. The first non-terminal symbol is by default the start symbol. As the BNF definition is a plug-in component of the system, it means that GE can produce code in any language thereby giving the system flexibility.

\subsection{Grammar Parsing}
\label{subsec:parsing_grammars}

Instead of a handwritten tokenization parser (as implemented in previous versions of GE \cite{libGE,ONeill:2008:GEVA,PonyGE} and in other systems such as ECJ \cite{ECJ}), BNF grammars in PonyGE2 are parsed using regular expressions. The use of regular expressions allows other researchers to integrate parsing BNF grammars easily in their EC systems. The regular expressions have originally been created by \cite{Forstenlechner:2017:GRAMMARS}.

The parser allows for the separation of productions onto multiple lines, Python-esque line commenting with `\#', as well as single quotations within double quotations and vice versa for terminals. This allows for the creation of `meta-grammars'.

\subsection{Variable ranges in grammars}
\label{subsec:variable_grammar_ranges}

A useful special case is available when writing grammars: a production can be given as:

\begin{center}
	\texttt{GE\_RANGE:4}
\end{center}

\noindent for example, and this will be replaced by a set of productions: 

\begin{center}
	\texttt{0 | 1 | 2 | 3}.
\end{center}

With \texttt{GE\_RANGE:dataset\_n\_vars}, the number of productions will be set by the number of columns in the dataset. Using grammar productions like the following, we can avoid hard-coding the number of independent variables, as illustrated in the grammar excerpt shown in Fig.~\ref{fig:GE_range}.
\begin{figure}[!ht]
	\begin{Verbatim}
                        <var>    ::= x[<varidx>]    
                        <varidx> ::= GE_RANGE:dataset_n_vars
    \end{Verbatim}
    \caption{Grammar excerpt showing use of \texttt{GE\_range}.}
    \label{fig:GE_range}
\end{figure}

Along with the fitness function, the grammar is one of the most problem-specific components of the PonyGE2 algorithm. The performance of PonyGE2 can be greatly affected by the grammar.

\section{Linear Genome Representation}
\label{sec:linear_representation}

Canonical Grammatical Evolution uses linear genomes (also called chromosomes) to encode genetic information \cite{ONeill:2003:GE}. These linear genomes are then mapped via the use of a formal BNF-style grammar to produce a phenotypic output. All individuals in PonyGE2 have an associated linear genome which can be used to exactly reproduce that individual.

\subsection{Genotype-Phenotype Mapping Process}
\label{subsec:linear_mapping}

The genotype is used to map the start symbol as defined in the Grammar onto terminals by reading codons to generate a corresponding integer value, from which an appropriate production rule is selected by using the Mod (or modulus) rule:

\begin{center}
	\texttt{Rule = c \% r}
\end{center}

\noindent where \texttt{c} is the codon integer value, and \texttt{r} is the number of rule choices for the current non-terminal symbol.

Consider the rule described in Fig. \ref{fig:non_terminal_mapping}. Given the non-terminal \texttt{<op>} which describes a set of mathematical operators that can be used, there are four production rules to select from. As can be seen, the choices are effectively labelled with integers counting from zero.
\begin{figure}[!ht]
	\centering
	\begin{Verbatim}
                                 <op> ::= +    (0)
                                        | -    (1)
                                        | *    (2)
                                        | /    (3)
	\end{Verbatim}
    \caption{Definition of a non-terminal \texttt{<op>} with four terminal production choices.}
    \label{fig:non_terminal_mapping}
\end{figure}

If we assume the codon being read produces the integer 6, then \texttt{6 \% 4 = 2} would select rule (2) \texttt{*}. Therefore, the non-terminal \texttt{<op>} is replaced with the terminal \texttt{*} in the derivation string. Each time a production rule has to be selected to transform a non-terminal, another codon is read. In this way the system traverses the genome.

The linear genotype-to-phenotype mapping process in PonyGE2 compiles a full derivation tree for the individual in question by default (this process is detailed in Section \ref{sec:tree_representation}). However, in certain configurations (such as when all variation operators operate on the linear genome), PonyGE2 has no need to maintain the full derivation trees of individuals during the course of an evolutionary run\footnote{Note that this excludes the initialisation of the initial population.}. In this case, a separate mapper is used which only generates numerical information on aspects of the derivation tree such as the overall maximum derivation tree depth and the number of nodes in the tree, resulting in a substantial reduction in the run-time of the algorithm. Thus, individuals mapped from a genome will have the same attributes as those generated from a derivation tree. 

\subsection{Tails and Wrapping}
\label{subsec:wrapping}

The `used' portion of the genome (i.e. the portion of the genome that directly maps to the phenotype) may not necessarily cover the entire length of the genome. The remaining unused portion of the genome is referred to as the `tail' of the genome. When initialising individuals by derivation tree-based methods such as Sensible initialisation \cite{Ryan:2003:Sensible} or Position Independent Grow \cite{Fagan:2016:PI}, a complete individual is generated with a complete genome (i.e. the number of used codons is equal to the length of the initial genome). A tail of randomly generated codons is then appended to the complete genome. Tails in PonyGE2 are initialised at 50\% of the length of the original genome, as per recommendations described in \cite{Nicolau:2012:Tails}. However, it must be noted that the use of linear genome operators means that these tails may become used (i.e. tails are not maintained subsequent to initialisation).

Even with the presence of tails, during the genotype-to-phenotype mapping process, it is possible  to run out of codons before the mapping process has terminated. In this case, a \emph{wrapping} operator can be applied which results in the mapping process re-reading the genome again from the start (i.e. wrapping past the end of the genome back to the beginning). As such, codons are reused when wrapping occurs. This means that it is possible for  codons to be used two or more times depending on the number of wraps specified. GE works with or without wrapping, and wrapping has been shown to be useful on some problems \cite{ONeill:2003:GE}, however, it does come at the cost of introducing functional dependencies between codons that would not otherwise arise \cite{Nicolau:2012:Tails}.

By default, wrapping in PonyGE2 is not used, however it is possible to specify the desired maximum number of times the mapping process is permitted to wrap past the end of the genome back to the beginning again. Note that permitting the mapping process to wrap on genomes does not necessarily mean it will wrap across genomes. The provision is merely allowed.

\subsection{Invalid Individuals}
\label{subsec:invalids}

In GE each time the same codon is expressed it will always generate the same integer value, but depending on the current non-terminal to which it is being applied, it may result in the selection of a different production rule. This feature is referred to as ``intrinsic polymorphism''. What is crucial however, is that each time a particular individual is mapped from its genotype to its phenotype, the same output is generated. This is the case because the same choices are made each time. In some cases it is possible that an incomplete mapping could occur; if the genome has been completely traversed (even after multiple wrapping events), and the derivation string (i.e. the derived expression) still contains non-terminals, such an individual is dubbed \emph{invalid} as it will never undergo a complete mapping to a set of terminals. For this reason an upper limit on the number of wrapping events that can occur is imposed (as detailed in Section \ref{subsec:wrapping}), otherwise mapping could continue indefinitely in this case. In the case of an invalid individual, the mapping process is typically aborted and the individual in question is given the lowest possible fitness value. The selection and replacement mechanisms then operate accordingly to increase the likelihood that this individual is removed from the population.

To reduce the number of invalid individuals being passed from generation to generation various strategies can be employed. Strong selection pressure could be applied, for example, through a steady state replacement. Alternatively, a repair strategy can be adopted which ensures that every individual results in a valid program. For example, in the case that there are non-terminals remaining after using all the genetic material of an individual (with or without the use of wrapping) default rules for each non-terminal can be pre-specified that are used to complete the mapping in a deterministic fashion. Another strategy is to remove the recursive production rules that cause an individual’s phenotype to grow, and then to reuse the genotype to select from the remaining non-recursive rules. Finally, the use of genetic operators which manipulate the derivation tree rather than the linear genome can be used to ensure the generation of completely mapped phenotype strings.

\subsection{A note on unit productions}
\label{subsec:unit_productions}

A {\em unit production} is a production which is the only production on the right-hand side of a rule. Traditionally, GE would not consume a codon for unit productions. This was a design decision taken by O'Neill et al.~\cite{ONeill:2003:GE}. However, in PonyGE2 unit productions consume codons, the logic being that it helps to do linear tree-style operations. 

The original design decision on unit productions was also taken before the introduction of evolvable grammars whereby the arity of a unit production could change over time. In this case consuming codons will help to limit the ripple effect from that change in arity. 

In summary, the merits for not consuming a codon for unit productions are not clearly defined in the literature. The benefits in consuming codons are a reduction in computation and improved speed with linear tree style operations. Other benefits are an increase in non-coding regions in the chromosome that through evolution of the grammar may then express useful information. 

\section{Derivation Tree Representation}
\label{sec:tree_representation}

During the linear genotype-to-phenotype mapping process, a derivation tree is implicitly generated; since each production choice generates a codon, it can be viewed as a node in an overall derivation tree. The parent rule that generated that choice is viewed as the parent node, and any production choices resultant from non-terminals in the current production choice are viewed as child nodes. The depth of a particular node is defined as how many parents exist in the tree directly above it, with the root node of the entire tree (the start symbol of the grammar) being at depth 1. Finally, the root of each individual node in the derivation tree is the non-terminal production rule that generated the node choice itself. A full derivation tree of a PonyGE2 individual is encoded as a recursive class, with all nodes in the tree being instances of that class.

While linear genome mapping means that each individual codon specifies the production choice to be selected from the given production rule, it is possible to do the opposite. Deriving an individual solution purely using the derivation tree (i.e. \emph{not} using the genotype-to-phenotype mapping process defined in Section \ref{subsec:linear_mapping}) is entirely possible, and indeed provides a lot more flexibility towards the generation of individuals than a linear mapping.

In a derivation tree based mapping process, each individual begins with the start rule of the grammar (as with the linear mapping). However, instead of a codon from the genome defining the production to be chosen from the given rule, a random production is chosen. Once a production is chosen, it is then possible to retroactively {\em create} a codon that would result in that same production being chosen if a linear mapping were to be used. In order to generate a viable codon, first the index of the chosen production is taken from the overall list of production choices for that rule. Then, a random integer from within the range:

\begin{center}
	\texttt{[no.~choices : no.~choices : CODON\_SIZE]}
\end{center}

\noindent (i.e. a number from \texttt{no.~choices} to \texttt{CODON\_SIZE} with a step size of \texttt{no.~choices}). Finally, the index of the chosen production is added to this random integer. This results in a codon which will re-produce the production choice. For example, consider the following rule:

\begin{center}
	\texttt{<e> ::= a | b | c}
\end{center}

Now, let us randomly select the production choice \texttt{b}. The index of production choice \texttt{b} is 1. Next, we randomly select an integer from within the range \texttt{[3: 3: CODON\_SIZE]}, giving us a random number of 768. Finally, we add the index of production choice \texttt{b}, to give a codon of 769. In this manner it is possible to build a derivation tree, where each node will have an associated codon. Simply combining all codons into a list gives the full genome for the individual.

Importantly, since the genome does not define the mapping process, invalid solutions can not be generated by derivation tree-based methods.

\subsection{Context-Aware Operations}

Since production choices are not set with the use of a derivation tree representation (i.e. the production choice defines the codon, rather than the codon defining the production choice), it is possible to build derivation trees in an intelligent manner by restricting certain production choices. For example, it is possible to force derivation trees to a certain depth by only allowing recursive production choices to be made until the tree is deep enough that branches can be terminated at the desired depth. This is the basis of context-aware derivation methods such as Ramped Half-and-Half (or Sensible) initialisation \cite{Ryan:2003:Sensible}.

It is also possible to perform intelligent variation operations using derivation tree methods. For example, crossover and mutation can be controlled by only selecting specific types of sub-trees for variation (e.g. sub-trees of specific sizes or sub-trees rooted at specific nodes). Note that the use of derivation tree-based operators comes at the expense of increased computational run-time.

In general, the use of a linear genome does not allow for such context-aware operations, i.e. operations on linear genomes are performed randomly, without reference to the effect or output of any particular portion of the genome. Although intelligent linear genome operators exist, e.g. \cite{Byrne:2009:Mutation}, they are not implemented in PonyGE2 as similar functions can be performed in a simpler manner using derivation-tree based operations.

\section{Operators}
\label{sec:operators}

This section contains a list of all operators currently implemented in PonyGE2.

\subsection{Initialisation}
\label{subsec:initialisation}

There are two main ways to initialise a GE individual: by generating a genome, or by generating a derivation tree. Generation of a genome can only be done by creating a random genome string, and as such the use of genome initialisation cannot guarantee control over any aspects of the initial population. Population initialisation via derivation tree generation on the other hand allows for fine control over many aspects of the initial population, e.g. depth limits or derivation tree shape. Unlike with genome initialisation, there are a number of different ways to initialise a population using derivation trees. Currently implemented methods are detailed below.

\textbf{\subsubsection{Linear genome initialisation}~}

At present, the only method for initialising a population of individuals through the use of linear genomes in Grammatical Evolution is to generate random genome strings, known as Random Genome Initialisation. Random genome initialisation in Grammatical Evolution should be used with caution as poor grammar design can have a negative impact on the quality of randomly initialised solutions due to the inherent bias capabilities of GE \cite{Fagan:2016:PI,Nicolau:2016:Repetition}.

\textbf{\subsubsection{Derivation tree initialisation}~}

Initialising a population of individuals through the use of derivation tree-based methods allows for much greater control over many aspects of individuals in the population, including derivation tree depth, number of nodes, and shape. At present, there are three such initialisation methods in PonyGE2, outlined below.

\smallskip

\textbf{Random tree initialisation}
            
\noindent Random derivation tree initialisation generates individuals by randomly building derivation trees up to the specified maximum initialisation depth limit. This is analogous to using the \texttt{Grow} component of Ramped Half-and-Half/Sensible initialisation to generate an entire population \cite{Ryan:2003:Sensible}. Note that there is no obligation that randomly generated derivation trees will extend to the depth limit; they will be of random size, but depending on how the grammar is written they may have a tendency towards smaller tree sizes with the use of a grammar-based mapping \cite{Fagan:2016:PI,Nicolau:2016:Repetition}.

\smallskip

\textbf{Ramped Half-and-Half/Sensible Initialisation} \cite{Ryan:2003:Sensible}
            
\noindent Ramped Half-and-Half initialisation in Grammatical Evolution is often called ``Sensible Initialisation" \cite{Ryan:2003:Sensible}. Sensible Initialisation follows traditional GP Ramped Half-and-Half initialisation by initialising a population of individuals using two separate methods: \texttt{Full} and \texttt{Grow}. \texttt{Full} initialisation generates a derivation tree where all branches extend to the specified depth limit. This tends to generate very bushy, evenly balanced trees \cite{Fagan:2016:PI}. \texttt{Grow} initialisation generates a randomly built derivation tree where no branch extends past the depth limit. 

Note that the \texttt{Grow} component of Sensible initialisation is analogous to random derivation tree initialisation, i.e. no branch in the tree is \emph{forced} to reach the specified depth. Depending on how the grammar is written, this can result in a very high probability of small trees being generated, regardless of the specified depth limit \cite{Fagan:2016:PI}. Note also that RHH initialisation with the use of a grammar-based mapping process such as GE can potentially result in a high number of duplicate individuals in the initial generation, resulting from a potentially high number of very small solutions \cite{Fagan:2016:PI,Harper:2010:Initialisation,Nicolau:2016:Repetition}. As such, caution is advised when using RHH initialisation in grammar-based systems, as particular care needs to be given to grammar design in order to minimise this effect \cite{Fagan:2016:PI, Harper:2010:Initialisation}.

\smallskip

\textbf{Position Independent Grow Initialisation} \cite{Fagan:2016:PI}
            
\noindent Position Independent Grow (PI Grow) initialisation in Grammatical Evolution mirrors Sensible/Ramped Half-and-Half initialisation by initialising a population of individuals over a ramped range of depths. However, while RHH uses two separate methods \texttt{Full} and \texttt{Grow} to generate pairs of individuals at each depth, PI Grow eschews the \texttt{Full} component and only uses the \texttt{Grow} aspect. There are two further differences between traditional GP \texttt{Grow} and PI Grow \cite{Fagan:2016:PI}:

	\begin{enumerate}
		\item At least one branch of the derivation tree is forced to the specified maximum depth in PI Grow, and
		\item Non-terminals are expanded in random (i.e. position independent) order rather than the left-first derivation of traditional mappers.
	\end{enumerate}

\subsection{Selection}
\label{subsec:selection}

The selection operator takes the original Generation $n$ population and produces a parent population to be used by the variation operators. As detailed in Section \ref{subsec:invalids}, the linear genome mapping process in Grammatical Evolution can generate invalid individuals. Only valid individuals are selected by default in PonyGE2, however this can be changed with the use of an optional argument.

Two selection operators are provided in PonyGE2. These operators are detailed below.

\textbf{\subsubsection{Tournament Selection}~}
\label{subsubsec:tournament}

Tournament selection randomly selects \texttt{tournament\_size} individuals from the overall population and returns the best. This process continues until \texttt{generation\_size} individuals have been selected. If no elitism is used, the \texttt{generation\_size} is equal to the full \texttt{population\_size}. However, if elitism is used, the \texttt{generation\_size} is equal to the full \texttt{population\_size} minus the number of elites. This prevents extra individuals from being generated and evaluated which would constitute additional search.

\textbf{\subsubsection{Truncation Selection}~}

Truncation selection takes an entire population, sorts it, and returns a specified top proportion of that population.

\subsection{Variation}
\label{subsec:variation}

Variation operators in evolutionary algorithms explore the search space by varying genetic material of individuals in order to explore new areas of the search space. The two main types of variation operator implemented in PonyGE2 are Crossover and Mutation.

\textbf{\subsubsection{Crossover}~}
        
Crossover randomly selects pairs of parents from the parent population created by the selection process. Unlike canonical Genetic Programming \cite{Koza:1992:GP}, crossover in Grammatical Evolution always produces \emph{two} children from these two parents \cite{ONeill:2003:Crossover}. As with Tournament Selection, Crossover in PonyGE2 continues until \texttt{generation\_size} children have been generated (i.e. crossover operates over the entire parent population rather than a specified percentage of that population).

One derivation tree-based crossover operator is provided in PonyGE2, along with four linear crossover operators. Note that with all linear genome crossovers, crossover points are selected within the used portion of the genome by default (i.e. crossover does not occur in the unused tail of the individual). Note also that while subtree-based operators do not allow invalid individuals to be generated, this is possible with all linear operators.

\smallskip

\textbf{Fixed Onepoint Crossover}

\noindent Given two individuals, fixed onepoint crossover creates two children by selecting the same point on both genomes for crossover to occur. The head of genome 0 is then combined with the tail of genome 1, and the head of genome 1 is combined with the tail of genome 0. This means that genomes will always remain the same length after crossover. 

\smallskip

\textbf{Fixed Twopoint Crossover}

\noindent Given two individuals, fixed twopoint crossover creates two children by selecting the same points on both genomes for crossover to occur. The head and tail of genome 0 are then combined with the mid-section of genome 1, and the head and tail of genome 1 are combined with the mid-section of genome 0. This means that genomes will always remain the same length after crossover.

\smallskip

\textbf{Variable Onepoint Crossover}

\noindent Given two individuals, variable onepoint crossover creates two children by selecting a different point on each genome for crossover to occur. The head of genome 0 is then combined with the tail of genome 1, and the head of genome 1 is combined with the tail of genome 0. This allows genomes to grow or shrink in length.

\smallskip

\textbf{Variable Twopoint Crossover}

\noindent Given two individuals, variable twopoint crossover creates two children by selecting two different points on each genome for crossover to occur. The head and tail of genome 0 are then combined with the mid-section of genome 1, and the head and tail of genome 1 are combined with the mid-section of genome 0. This allows genomes to grow or shrink in length.

\textbf{\subsubsection{Mutation}~}
        
While crossover operates on pairs of selected parents to produce new children, mutation in Grammatical Evolution operates on every individual in the child population \emph{after} crossover has been applied. Note that this is different in implementation so canonical GP crossover and mutation, whereby a certain percentage of the population would be selected for crossover with the remaining members of the population subjected to mutation \cite{Koza:1992:GP}.

One subtree mutation operator is provided in PonyGE2, along with to linear genome mutation operators, detailed below. By default, linear genome mutation operators in PonyGE2 operate only on the used portion of the genome.

\smallskip

\textbf{Codon-based Integer Flip Mutation}

\noindent Codon-based integer flip mutation randomly mutates every individual codon in the genome with a certain probability.

\smallskip

\textbf{Genome-based Integer Flip Mutation}

\noindent Genome-based integer flip mutation mutates a specified number of codons randomly selected from the genome.

\subsection{Evaluation}
\label{subsec:evaluation}

PonyGE2 takes advantage of vectorised evaluation to enable fast evaluation on large dataset arrays for supervised learning problems. Furthermore, caching is provided in PonyGE2, along with a few options for dealing with cached individuals as discussed in \cite{Nicolau:2016:Repetition}. Multicore evaluation is also provided, but this feature is not currently supported on machines using a Windows OS.

\subsection{Replacement}
\label{subsec:replacement}

The replacement strategy for an Evolutionary Algorithm defines which parents and children survive into the next generation. Two replacement operators are provided in PonyGE2.

\textbf{\subsubsection{Generational Replacement with Elitism}~} 

Generational replacement replaces the entire parent population with the newly generated child population at every generation. Generational replacement is most commonly used in conjunction with elitism. With elitism, the best \texttt{ELITE\_SIZE} individuals in the parent population are copied over unchanged to the next generation. Elitism ensures continuity of the best ever solution at all stages through the evolutionary process, and allows for the best solution to be updated at each generation.

\textbf{\subsubsection{Steady State Replacement}~}

Steady state replacement uses the GENITOR model \cite{Whitley:1989:GENITOR} whereby new individuals directly replace the worst individuals in the population regardless of whether or not the new individuals are fitter than those they replace. Note that traditional GP crossover generates only 1 child \cite{Koza:1992:GP}, whereas linear GE crossover (and thus all crossover functions used in PonyGE2) generates 2 children from 2 parents~\cite{ONeill:2003:GE,ONeill:2003:Crossover}. Thus, PonyGE2 uses a deletion strategy of 2.

\section{Example Problems}
\label{sec:examples}

Four example problems are provided in the initial release of PonyGE2. These problems are described in this section.

\subsection{String-match}

The grammar specifies words as lists of vowels and consonants along with special characters. The aim is to match a target string. The default string match target is \texttt{Hello world!}.

\subsection{Regression}

The grammar generates a symbolic function composed of standard mathematical operations and a set of variables. This function is then evaluated using a pre-defined set of inputs, given in the datasets folder. Each problem suite has a unique set of inputs. The aim is to minimise some error between the expected output of the function and the desired output specified in the datasets. This is the default problem for PonyGE. The default dataset is the Vladislavleva-4 dataset~\cite{Vladislavleva:2009:Non_Linearity}.

\subsection{Classification}

Classification can be considered a special case of symbolic regression but with a different error metric. Like with regression, the grammar generates a symbolic function composed of standard mathematical operations and a set of variables. This function is then evaluated using a pre-defined set of inputs, given in the datasets folder. Each problem suite has a unique set of inputs. The aim is to minimise some classification error between the expected output of the function and the desired output specified in the datasets.

\subsection{Pymax}

One of the strongest aspects of a grammatical mapping approach such as PonyGE2 is the ability to generate executable computer programs in an arbitrary language \cite{ONeill:2003:GE}. In order to demonstrate this in the simplest way possible, we have included an example python programming problem.
 
The \texttt{Pymax} problem is a traditional maximisation problem, where the goal is to produce as large a number as possible. However, instead of encoding the grammar in a symbolic manner and evaluating the result, we have encoded the grammar for the \texttt{Pymax} problem as a basic Python programming example. The phenotypes generated by this grammar are executable python functions, whose outputs represent the fitness value of the individual. Users are encouraged to examine the \texttt{pymax.bnf} grammar, the \texttt{pymax.py} fitness function, and the resultant individual phenotypes to gain an understanding of how grammars can be used to generate such arbitrary programs~\cite{PonyGE2}.

%

\subsection{Adding New Problems}

It has been made as simple as possible to add new problems to PonyGE. To add a new problem, any number of the following may be required:

\begin{enumerate}
	\item a new grammar file named with a {\tt .bnf} suffix and placed in {\tt grammars/};
	\item a new fitness function implemented as a class in a file {\tt fitness/x.py} where {\tt x} is the name of the class (note that existing fitness functions may be re-used, e.g. for supervised learning problems);
	\item for supervised learning, a new dataset split into {\tt datasets/x/Train.csv} and {\tt datasets/x/Test.csv} where {\tt x} is a subdirectory named after the dataset.
\end{enumerate}

\section{Conclusions and Future Work}
\label{sec:future_work}

This paper described PonyGE2, a modern Python implementation of Grammatical Evolution. While this paper presents a brief overview of the system, comprehensive documentation is available on GitHub at \url{https://github.com/jmmcd/PonyGE2}. The codebase is fully commented to facilitate understanding and to provide ease of extensibility, and is PEP-8 compliant for readability. We welcome future contributors and collaborators from the wider field, and GitHub provides a forum for future discussion \cite{PonyGE2}.

A number of additions to PonyGE2 are planned in the immediate future. Development is ongoing, and will see the implementation of a number of additional features, including:

\begin{enumerate}
	\item Multi-objective optimisation using NSGA-II \cite{Deb:2002:NSGAII},
    \item Python packaging integration (e.g.~\texttt{setup.py}, \texttt{MANIFEST.in}, etc.): the aim is to have PonyGE2 PIP-installable.
    \item Parametrisable termination conditions,
    \item Extension of multicore evaluation support to Windows OS machines, and look into the integration of cloud based multicore support.
    \item Addition of more search engines and problems.
\end{enumerate}

Finally, PonyGE2 will be kept up to date with the most current best-of-practice techniques.

\section*{Acknowledgments}

This research is based upon works supported by Science Foundation Ireland under grant 13/IA/1850.

\bibliographystyle{abbrv}
\bibliography{sources} 

\begin{thebibliography}{10}

\bibitem{Byrne:2009:Mutation}
J.~Byrne, M.~O'Neill, and A.~Brabazon.
\newblock Structural and nodal mutation in grammatical evolution.
\newblock In {\em Proceedings of the 11th Annual conference on Genetic and
  evolutionary computation}, pages 1881--1882. ACM, 2009.

\bibitem{Deb:2002:NSGAII}
K.~Deb, A.~Pratap, S.~Agarwal, and T.~Meyarivan.
\newblock A fast and elitist multiobjective genetic algorithm: Nsga-ii.
\newblock {\em IEEE Transactions on Evolutionary Computation}, 6(2):182--197,
  2002.

\bibitem{Fagan:2016:PI}
D.~Fagan, M.~Fenton, and M.~O'Neill.
\newblock Exploring position independent initialisation in grammatical
  evolution.
\newblock In {\em Evolutionary Computation (CEC), 2016 IEEE Congress on}, pages
  5060--5067. IEEE, 2016.

\bibitem{PonyGE2}
M.~Fenton, J.~McDermott, D.~Fagan, E.~Hemberg, S.~Forstenlechner, and
  M.~O'Neill.
\newblock Ponyge2.
\newblock \url{https://github.com/jmmcd/PonyGE2}, 2017.

\bibitem{Forstenlechner:2017:GRAMMARS}
S.~Forstenlechner, D.~Fagan, M.~Nicolau, and M.~O'Neill.
\newblock A grammar design pattern for arbitrary program synthesis problems in
  genetic programming.
\newblock In {\em EuroGP 2017: Proceedings of the 20th European Conference on
  Genetic Programming}, LNCS, Amsterdam, Netherlands, 2017. Springer Verlag.
\newblock (forthcoming).

\bibitem{Harper:2010:Initialisation}
R.~Harper.
\newblock Ge, explosive grammars and the lasting legacy of bad initialisation.
\newblock In {\em Evolutionary Computation (CEC), 2010 IEEE Congress on}, pages
  1--8. IEEE, 2010.

\bibitem{Koza:1992:GP}
J.~R. Koza.
\newblock {\em Genetic programming: on the programming of computers by means of
  natural selection}, volume~1.
\newblock MIT press, 1992.

\bibitem{ECJ}
S.~Luke, L.~Panait, G.~Balan, S.~Paus, Z.~Skolicki, R.~Kicinger, E.~Popovici,
  K.~Sullivan, J.~Harrison, J.~Bassett, R.~Hubley, A.~Desai, A.~Chircop,
  J.~Compton, W.~Haddon, S.~Donnelly, B.~Jamil, J.~Zelibor, E.~Kangas,
  F.~Abidi, H.~Mooers, J.~O'Beirne, K.~A. Talukder, S.~McKay, and
  J.~McDermott".
\newblock Ecj, 2015.
\newblock V. 23.

\bibitem{PonyGE}
J.~McDermott and E.~Hemberg.
\newblock Ponyge.
\newblock \url{https://github.com/jmmcd/ponyge}, 2009.

\bibitem{McDermott:2012:Benchmarks}
J.~McDermott, D.~R. White, S.~Luke, L.~Manzoni, M.~Castelli, L.~Vanneschi,
  W.~Jaskowski, K.~Krawiec, R.~Harper, K.~De~Jong, et~al.
\newblock Genetic programming needs better benchmarks.
\newblock In {\em Proceedings of the 14th annual conference on Genetic and
  evolutionary computation}, pages 791--798. ACM, 2012.

\bibitem{Nicolau:2015:Guidelines}
M.~Nicolau, A.~Agapitos, M.~O'Neill, and A.~Brabazon.
\newblock Guidelines for defining benchmark problems in genetic programming.
\newblock In {\em Evolutionary Computation (CEC), 2015 IEEE Congress on}, pages
  1152--1159. IEEE, 2015.

\bibitem{Nicolau:2016:Repetition}
M.~Nicolau and M.~Fenton.
\newblock Managing repetition in grammar-based genetic programming.
\newblock In {\em Proceedings of the 2016 on Genetic and Evolutionary
  Computation Conference}, pages 765--772. ACM, 2016.

\bibitem{Nicolau:2012:Tails}
M.~Nicolau, M.~O'Neill, and A.~Brabazon.
\newblock Termination in grammatical evolution: Grammar design, wrapping, and
  tails.
\newblock In {\em Evolutionary Computation (CEC), 2012 IEEE Congress on}, pages
  1--8. IEEE, 2012.

\bibitem{libGE}
M.~Nicolau and D.~Slattery.
\newblock libge, 2006.

\bibitem{Noorian:2015:GE_R}
F.~Noorian, A.~M. de~Silva, and P.~H. Leong.
\newblock gramevol: Grammatical evolution in r.
\newblock {\em Journal of Statistical Software}, 2015.

\bibitem{ONeill:2008:GEVA}
M.~O'Neill, E.~Hemberg, C.~Gilligan, E.~Bartley, J.~McDermott, and A.~Brabazon.
\newblock Geva: grammatical evolution in java.
\newblock {\em ACM SIGEVOlution}, 3(2):17--22, 2008.

\bibitem{ONeill:2003:Crossover}
M.~O'Neill, C.~Ryan, M.~Keijzer, and M.~Cattolico.
\newblock Crossover in grammatical evolution.
\newblock {\em Genetic Programming and Evolvable Machines}, 4(1):67--93, 2003.

\bibitem{ONeill:2003:GE}
M.~O’Neill and C.~Ryan.
\newblock Grammatical evolution: Evolutionary automatic programming in a
  arbitrary language, 2003.

\bibitem{Ryan:2003:Sensible}
C.~Ryan and R.~M.~A. Azad.
\newblock Sensible initialisation in grammatical evolution.
\newblock In {\em GECCO}, pages 142--145, 2003.

\bibitem{GERET}
P.~Schumann.
\newblock Geret.
\newblock \url{http://www.geret.org/}, 2009.

\bibitem{PEP-8}
G.~van Rossum, B.~Warsaw, and N.~Coghlan.
\newblock Pep 8--style guide for python code, 2001.

\bibitem{Vladislavleva:2009:Non_Linearity}
E.~J. Vladislavleva, G.~F. Smits, and D.~den Hertog.
\newblock {Order of NonLinearity as a Complexity Measure for Models Generated
  by Symbolic Regression via Pareto Genetic Programming}.
\newblock {\em IEEE Transactions on Evolutionary Computation}, 13(2):333--349,
  2009.

\bibitem{Whigham:1995:GBGP}
P.~A. Whigham.
\newblock Grammatically-based genetic programming.
\newblock In {\em Proceedings of the workshop on genetic programming: from
  theory to real-world applications}, volume~16, pages 33--41, 1995.

\bibitem{Whitley:1989:GENITOR}
D.~Whitley.
\newblock The genitor algorithm and selection pressure: Why rank-based
  allocation of reproductive trials is best.
\newblock In {\em ICGA}, pages 116--123, 1989.

\end{thebibliography}

\end{document}